\pdfoutput=1
\documentclass[11pt]{article}
\usepackage[final]{acl}

\usepackage{times}
\usepackage{latexsym}

\usepackage[T1]{fontenc}
\usepackage[utf8]{inputenc}
\usepackage{microtype}
\usepackage{inconsolata}
\usepackage{graphicx}
\usepackage{tabularx}
\usepackage{xcolor}
\usepackage{booktabs}
\usepackage{amsmath}
\usepackage{subcaption}
\usepackage{float}
\usepackage{placeins}
\usepackage{fancyvrb}
\usepackage{fvextra} 
\usepackage{pgfplots}
\usepackage{url}
\pgfplotsset{compat=1.18}
\usepackage{tikz}
\DefineVerbatimEnvironment{PromptVerb}{Verbatim}{
  fontsize=\small,
  breaklines,     
  breaksymbol={},    
  breaksymbolindent=0pt
}

\title{Understanding Multi-Agent Reasoning with Large Language Models for Cartoon VQA}

\author{
\normalfont
\begin{tabular}{c c}
Tong Wu & Thanet Markchom \\
\texttt{tongwuwhitney@gmail.com} &
University of Reading \\
& Reading, UK \\
& \texttt{thanet.markchom@reading.ac.uk}
\end{tabular}
}

\begin{document}
\maketitle
\begin{abstract}
Visual Question Answering (VQA) for stylised cartoon imagery presents challenges, such as interpreting exaggerated visual abstraction and narrative-driven context, which are not adequately addressed by standard large language models (LLMs) trained on natural images. To investigate this issue, a multi-agent LLM framework is introduced, specifically designed for VQA tasks in cartoon imagery. The proposed architecture consists of three specialised agents: visual agent, language agent and critic agent, which work collaboratively to support structured reasoning by integrating visual cues and narrative context. The framework was systematically evaluated on two cartoon-based VQA datasets: Pororo and Simpsons. Experimental results provide a detailed analysis of how each agent contributes to the final prediction, offering a deeper understanding of LLM-based multi-agent behaviour in cartoon VQA and multimodal inference. The source code is available at \url{https://github.com/tongwu17/Multi_Agent_Cartoon}.
\end{abstract}

\section{Introduction}
VQA has emerged as a benchmark task for evaluating models’ visual reasoning capabilities. These benchmarks, however, lack the abstraction, stylisation, and narrative richness present in cartoons and animated scenes, whose linguistic and visual expressions often fall outside the natural-language distributions that contemporary LLMs are primarily trained on.

Recent advancements in LLMs, particularly multimodal variants \cite{liu2024multimodal}, have shown promise in combining image and textual understanding. However, existing applications largely overlook stylised visual inputs like those found in cartoons, which often require contextual inference, visual reasoning, and narrative tracking.

The study investigates how a multi-agent reasoning framework behaves on cartoon-based VQA tasks and what contributions its individual agents make. The framework decomposes the problem into three subtasks - visual description, language answering, and critic detection - each handled by a dedicated agent prompted via a foundation LLM. Based on experiments on two distinct cartoon VQA datasets, Pororo and Simpsons, the analysis reveals when and how these agents collaborate or fail, thereby providing a deeper understanding of LLM-based multi-agent contributions in character-driven and satire-rich cartoon domains. The main contributions of this work are:
\begin{list}{\textbullet}{%
  \setlength{\leftmargin}{1.0em}
  \setlength{\labelwidth}{0.6em}
  \setlength{\labelsep}{0.35em}
  \setlength{\itemindent}{0pt}
  \setlength{\itemsep}{0.2em}
  \setlength{\parsep}{0pt}
  \setlength{\topsep}{0.15em}
  \setlength{\partopsep}{0pt}
}
\item A systematic multi-agent framework is established to analyse how three specialised agents (visual, language, and critic) behave in stylised cartoon VQA, with a focus on their individual and joint contributions.
\item This is the first work to examine how different agent combinations influence answer quality and narrative alignment in cartoon VQA.
\item A comparison between vision encoders (BLIP-2-based visual features) and the generative LLM GPT-4o-mini is conducted to analyse how explicit visual grounding affects performance relative to language-only prompting and its interaction with multi-agent configurations.
\end{list}

\section{Related Work}
\subsection{VQA Research Landscape}
VQA has been extensively studied in the context of natural images, where models are tasked with answering questions based on real-world visual content. Datasets such as GQA \cite{Hudson_2019_CVPR}, VizWiz \cite{VizWiz_2018}, and the 360-degree images \cite{Chou_2020_WACV} all capture everyday scenes from realistic visual perspectives. These benchmarks emphasise factual reasoning, spatial understanding, and object recognition grounded in natural imagery.

Beyond natural images, VQA research has expanded to abstract, symbolic, and domain-specific visual content. DVQA \cite{kafle2018} targets reasoning over charts in data visualisations, DocVQA \cite{mathew2021} focuses on textual and layout understanding in document images, while ArtVQA \cite{garcia2020AQUA} involves the interpretation of paintings.

\subsection{Cartoon and Cartoon-like Visual Understanding}
Cartoon imagery poses unique challenges for vision-language models due to its exaggerated abstraction, symbolic visual elements, and episodic storytelling structure. These characteristics introduce reasoning demands that differ significantly from those in natural images, including symbolic interpretation, non-literal spatial layouts, and narrative tracking across frames.

Some work has directly addressed cartoon imagery. For instance, the Fahad18 dataset \cite{fahad18_2012} was developed for color-based cartoon character detection. In parallel, stylised media such as comics and manga have been largely explored for foundational computer vision tasks, including panel segmentation and bubble detection. Datasets such as Manga109 \cite{Matsui_2016}, eBDtheque \cite{ebdtheque2013}, and DCM772 \cite{dcm_2018} have supported applications such as panel segmentation, speech bubble extraction, and character face detection. More recently, LLMs have been applied to generation and comprehension tasks in these domains. For example, \citet{wu2025diffsenseibridgingmultimodalllms} propose using LLMs for controllable comic panel generation, while \citet{ikuta2024mangaubmangaunderstandingbenchmark} introduce a benchmark to evaluate LLMs’ understanding of comic narratives.

Cartoon-based VQA datasets are relatively scarce. Among the few available resources, PororoQA \cite{ijcai2017p280} is derived from the children’s cartoon video series Pororo, offering scene-level annotations and dialogues. SimpsonsVQA \cite{huynh2024simpsonsvqaenhancinginquirybasedlearning} constructs a VQA dataset based on The Simpsons cartoon. CausalChaos \cite{parmar2024causalchaos} is built upon the iconic Tom and Jerry cartoon series, aiming to foster causal reasoning in video-based question answering.

\subsection{Multi-Agent Systems}
Two dominant approaches have emerged in the development of LLM-based multi-agent systems. The first approach involves assigning a distinct LLM to each agent. In this paradigm, different LLMs act as specialised agents based on their unique capabilities and strengths. For instance, MetaGPT \cite{hong2024metagpt} defines agents such as product managers, architects, and engineers by invoking different LLM APIs. This model-level heterogeneity supports flexible agent design and functional specialisation, although it may increase implementation complexity and cost. The second approach constructs multiple agents by applying task-specific prompts within a shared underlying LLM. This prompt-based differentiation enables functional role separation without modifying the model architecture.  For example, CAMEL \cite{li2023camel} employs two instances of GPT-3.5-turbo, simulating the roles of an AI assistant and a user. Similarly, CoMM \cite{comm2024} orchestrates multiple reasoning paths within a unified LLM through prompt-defined agent instructions. 

The Multi-Agent VQA framework \cite{jiang2024multiagentvqaexploringmultiagent} introduces dedicated agents as tools for exploring multi-agent capabilities in foundation models for VQA, marking an early step in this direction.

\section{Methodology} 
\begin{figure}[ht]
  \centering
  \includegraphics[width=0.9\linewidth]{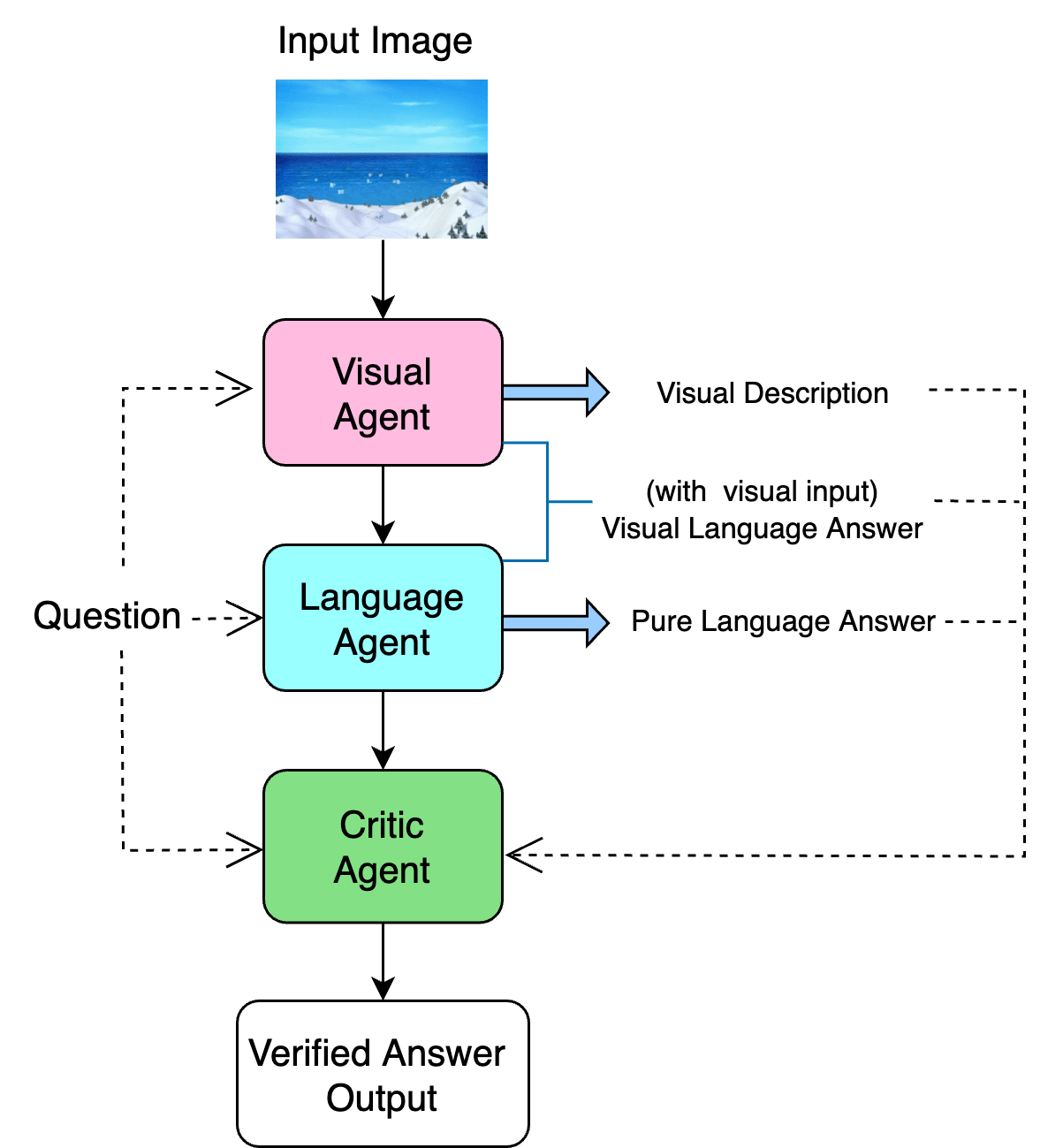}
  \caption{Overview of the multi-agent VQA pipeline}
  \label{fig:multi-agent-architecture}
\end{figure}

To address the limitations of single-agent vision-language models in handling stylised and narrative-rich cartoon imagery, a modular multi-agent question answering framework is proposed. The system decomposes the VQA task into three specialised stages: visual description, language answering, and critic checking. Each subtask is handled by a dedicated agent prompted through a vision-language foundation model, as illustrated in Figure~\ref{fig:multi-agent-architecture}.

\subsection{Visual Agent} 
The visual agent processes cartoon images to produce detailed scene descriptions. Cartoon-specific cues such as exaggerated expressions, artistic conventions and symbolic elements are captured through prompt engineering tailored for cartoon imagery. Each image (or GIF) is encoded and passed to a multimodal LLM (GPT-4o-mini) \cite{gpt4o2024}, which returns a textual description used as grounding for downstream reasoning.

As an additional point of comparison, a BLIP-2 based vision-language model \cite{pmlr-v202-li23q} is also included. In this configuration, each image is processed by BLIP-2, which keeps the visual encoder frozen and produces short visual descriptions from the extracted features. These BLIP-2 outputs function as explicit visual context, enabling an examination of how externally generated visual representations differ from the end-to-end multimodal prompting of GPT-4o-mini.

\subsection{Language Agent}
The language agent generates an answer based on the original user question and available narrative cues. Prompts are designed to encourage concise, context-aware answers without speculative or explanatory content. This agent primarily focuses on capturing narrative coherence and character-driven reasoning within the given context.

\subsection{Critic Agent}
The critic agent verifies and refines candidate answers based on multimodal evidence. It evaluates whether the prediction is semantically valid and factually grounded by analysing the question type together with the associated visual description and any available textual context (e.g., scene descriptions or subtitles). The agent performs a multi-step process: (i) assessing the reliability of visual input, (ii) selecting appropriate reasoning strategies tailored to question types (e.g., colour/counting, object recognition, dialogue consistency, spatial reasoning), (iii) estimating a model-confidence score, and (iv) determining whether to preserve or revise the answer. The final output includes the revised answer (if applicable), along with an explanation, confidence score, and supporting visual rationale.

\section{Experiments}
All experiments were conducted under an open-ended VQA setting. In this framework, the model is required to generate free-form answers to each question without being provided with predefined correct answers. For every question, an LLM is prompted with the corresponding visual input and question text to produce a natural language response. The generated answer is then compared against a reference ground-truth answer using a soft accuracy metric and commonly used text-generation metrics. This evaluation protocol is designed to better reflect the model's reasoning and language generation capabilities in a more realistic and challenging scenario, particularly when multiple agents are involved in the answering process.

\paragraph{Prompt Availability}
Due to space constraints, all prompts are provided in the project repository.

\subsection{Datasets}
Two cartoon VQA datasets are used:\\
\textbf{Pororo VQA} \cite{ijcai2017p280} is a video-based QA dataset derived from the animated series \textit{Pororo}.  Subtitle information was used to construct dialogues without speaker annotations. Each video scene was converted into an animated GIF for annotation. Each question is associated with five candidate answers, one of which is marked as correct. In this study, the GIFs were extracted into static images and encoded in base64 format for input into the LLM. Only the Pororo\_ENGLISH1 subset was used in the experiments. A fixed random seed was applied to sample questions from this subset to ensure experimental reproducibility. To better evaluate the model’s understanding of both visual scenes and textual descriptions, only the correct answer from each question was retained as ground truth. The model was prompted to generate a free-form answer without access to the original candidate options, enabling a more realistic assessment of its generative reasoning capabilities.
    
\textbf{Simpsons VQA} \cite{huynh2024simpsonsvqaenhancinginquirybasedlearning} is a large-scale cartoon-based VQA dataset comprising 23,269 images extracted from the animated series \textit{The Simpsons}, along with 166,533 QA pairs covering a wide range of reasoning skills. Each image is associated with between 1 and 11 QA pairs (7.2 on average). Each QA pair was manually reviewed by qualified workers and assigned two separate scores: question relevance (1 for relevant, 0 for irrelevant) and answer correctness (0 for correct, 0.5 for ambiguous, and 0 for incorrect). Following the official dataset guidelines, only the validation split is used in this study, as the test set remains private and undisclosed. To ensure proper evaluation, only high-quality pairs, defined as those with an overall score of 1, were retained. A fixed-size subset was randomly sampled using a fixed random seed to maintain experimental reproducibility. The visual inputs were converted to base64-encoded static images and fed into the LLM alongside the question, prompting the model to generate a free-form response without access to answer choices.

\subsection{Evaluation Metrics} 
LLMs have been shown to be effective evaluators in open-ended generation tasks, achieving human-level agreement in multiple settings \cite{zheng2023judging}. Structured scoring scales, such as 5-point intervals, have been shown to improve consistency across evaluation tasks \cite{lee-etal-2025-evaluating}. Fine-grained scoring is also recommended to better handle uncertainty in human annotations \cite{elangovan2025beyond}.

In this study, a five-level scoring scheme $\{0.0, 0.25, 0.5, 0.75, 1.0\}$ is adopted. The highest score (1.0) indicates full semantic equivalence with the ground truth, while lower scores reflect increasing degrees of inaccuracy or irrelevance.

All scores are generated by prompting an LLM with the input question, predicted answer, ground-truth reference, and answer type. The final performance is reported as the average score:

\[
\text{Average Accuracy} = \frac{1}{N} \sum_{i=1}^{N} s_i
\]

Besides the prompt-based evaluation method, this work also adopted commonly used metrics for text generation tasks, including: BLEU, ROUGE, METEOR, BERTScore, and BLEURT. 

\textbf{BLEU} measures n-gram overlap between generated and reference texts, combined with a brevity penalty to discourage overly short outputs. In this work, BLEU-1, BLEU-2, and BLEU-3 (corresponding to uni-gram through four-gram overlap) were computed. \textbf{ROUGE} evaluates recall-oriented n-gram overlap between generated and reference texts. This work adopted ROUGE-1, ROUGE-2, and ROUGE-L for evaluation. ROUGE-1 and ROUGE-2 consider uni-gram and bi-gram, respectively, while ROUGE-L is based on the longest common subsequence.
\textbf{METEOR} aligns generated and reference texts using exact matches, stemmed forms, and synonym mappings. It also attends to word order, making it more sensitive to semantic similarity than BLEU and ROUGE.

\textbf{BERTScore}~\cite{zhang2019bertscore} computes similarity between generated and reference texts by aligning contextualised token embeddings from a pretrained language model. By operating in embedding space rather than exact word matches, it better captures paraphrasing and meaning preservation.
\textbf{BLEURT}~\cite{sellam2020bleurt} is a learned evaluation metric based on a pretrained language model (BLEURT-20) fine-tuned on human judgment data. It outputs a scalar score reflecting semantic adequacy, fluency, and quality, and has been shown to correlate more strongly with human evaluations than traditional lexical overlap metrics.

\subsection{Baseline} 
The baseline corresponds to a single-agent setting in which only the language agent is used. In this configuration, GPT-4o-mini receives the image (base64-encoded) together with the question and produces an answer directly, without the visual-agent description stage or the critic verification stage introduced in the proposed framework. This baseline represents the behaviour of a cartoon-aware multimodal LLM when no multi-agent decomposition is applied, and therefore serves as the primary point of comparison for evaluating the contributions of the additional agents.

\subsection{Ablation Study}
To quantify the contribution of each agent, an ablation study is conducted by selectively removing the visual and critic components from the full pipeline. Four configurations are evaluated: Full (Visual + Language + Critic), Language Only, Visual + Language (without Critic), and Language + Critic (without Visual). The same configurations are applied to both Pororo and Simpsons under the open-ended VQA setting described above. All configurations use the same backbone model and decoding settings to ensure a fair comparison; only the availability of intermediate descriptions and critic-based revision varies across setups.

\subsection{Visual Encoder Comparison with BLIP-2}
To further isolate the contribution of the cartoon-specific visual agent, an additional ablation study is conducted in which the visual descriptions generated by the GPT-4o-mini-based visual agent are replaced with those produced by an off-the-shelf BLIP2 visual encoder. For each image, a frozen BLIP2 model is used to generate a short caption describing the scene, and this caption is passed to the downstream language and critic agents in exactly the same format as the original visual-agent outputs. All prompts, decoding parameters, and critic procedures are kept fixed, so that any performance differences can be attributed to the change in the source of visual descriptions.

\section{Results and Discussion}
\subsection{Ablation Study of Different Agent Configurations}
This section analyses the ablation results on the Pororo and Simpsons datasets, focusing on how different agent configurations affect performance. The Language only configuration is treated as the single-agent baseline, while the Full configuration represents the proposed multi-agent framework. The results are summarised in Table~\ref{tab:ablation-all}.

\begin{table*}[t]
    \centering
    \caption{Ablation study on the Pororo and Simpsons under various configurations: Full (Visual + Language + Critic), Language Only, Visual + Language (without Critic), and Language + Critic (without Visual)}
    \label{tab:ablation-all}
    
    \begin{minipage}{\linewidth}
        \centering
        \caption*{(a) Ablation Study on the Pororo Dataset}
        \small  
        \resizebox{\linewidth}{!}{%
        \begin{tabular}{lllllllllll}
            \toprule
            \textbf{Config.} & \textbf{Accuracy} & \textbf{BLEU-1} & \textbf{BLEU-2} & \textbf{BLEU-3} & \textbf{ROUGE-1} & \textbf{ROUGE-2} & \textbf{ROUGE-L} & \textbf{METEOR} & \textbf{BERTScore} & \textbf{BLEURT} \\
            \midrule
            Full & 0.8375 & 0.3527 & 0.2642 & 0.1993 & 0.5256 & 0.3549 & 0.5001 & 0.4482 & 0.6593 & 0.5665 \\
            Language Only & 0.8187 (\(-2.24\%\)) & 0.3831 & 0.2896 & 0.2182 & 0.5415 & 0.3579 & 0.5158 & 0.4600 & 0.6640 & 0.5622 \\
            Visual + Language & 0.8313 (\(-0.74\%\)) & 0.3296 & 0.2432 & 0.1879 & 0.5000 & 0.3213 & 0.4670 & 0.4372 & 0.6525 & 0.5544 \\
            Language + Critic & 0.8250 (\(-1.49\%\)) & 0.3831 & 0.2896 & 0.2182 & 0.5415 & 0.3579 & 0.5158 & 0.4600 & 0.6640 & 0.5622 \\
            \bottomrule
        \end{tabular}}
    \end{minipage}

    \vspace{0.3em}  

    \begin{minipage}{\linewidth}
        \centering
        \caption*{(b) Ablation Study on the Simpsons Dataset}
        \small
        \resizebox{\linewidth}{!}{%
        \begin{tabular}{lllllllllll}
            \toprule
            \textbf{Config.} & \textbf{Accuracy} & \textbf{BLEU-1} & \textbf{BLEU-2} & \textbf{BLEU-3} & \textbf{ROUGE-1} & \textbf{ROUGE-2} & \textbf{ROUGE-L} & \textbf{METEOR} & \textbf{BERTScore} & \textbf{BLEURT} \\
            \midrule
            Full & 0.8819 & 0.7500 & 0.2372 & 0.1616 & 0.8056 & 0.0000 & 0.8056 & 0.4028 & 0.9270 & 0.5100 \\
            Language Only & 0.8403 (\(-4.72\%\)) & 0.7222 & 0.2284 & 0.1556 & 0.7778 & 0.0000 & 0.7778 & 0.3889 & 0.9235 & 0.4945 \\
            Visual + Language & 0.8819 (\(0\%\)) & 0.7500 & 0.2372 & 0.1616 & 0.8056 & 0.0000 & 0.8056 & 0.4028 & 0.9213 & 0.5099 \\
            Language + Critic & 0.8403 (\(-4.72\%\)) & 0.7222 & 0.2284 & 0.1556 & 0.7778 & 0.0000 & 0.7778 & 0.3889 & 0.9235 & 0.4945 \\
            \bottomrule
        \end{tabular}}
    \end{minipage}
\end{table*}

On \textbf{the Pororo dataset}, removing the visual agent and relying solely on the language agent led to a drop in accuracy of 2.24\%. However, in terms of lexical-based metrics, including BLEU, ROUGE, and METEOR, as well as the embedding-based metric BERTScore, the Language Only and Language + Critic configurations outperformed the Visual + Language configuration. This behaviour could be attributed to the tendency of the Visual + Language setting to generate longer answers containing additional details and paraphrased expressions. Such variations reduced n-gram overlap and embedding similarity with the reference answers, resulting in lower scores on lexical-based and embedding-based metrics. 

The Visual + Language configuration performed closer to the Full configuration in terms of accuracy, with only a minor reduction of 0.74\%. However, it showed a consistent degradation across other metrics. This suggests that visual grounding alone is insufficient to ensure both correctness and answer quality. The critic component plays a vital role in refining and validating the final output, improving alignment between visual understanding and textual reasoning. This observation is further supported by the Language + Critic configuration, which improves over Language Only in accuracy.

On \textbf{the Simpsons dataset}, the performance gap between configurations was more pronounced. The Full and Visual + Language configurations achieved identical accuracy. Similarly, the Language-only and Language + Critic configurations also achieved identical accuracy. This suggests that, for this dataset, the critic agent had no impact on altering the answers generated by the language agent. This may be due to the fact that the answers in this dataset are typically single words. As a result, the critic did not play a similar role in removing unnecessary details from the answers or modifying them to improve accuracy, as observed in the Pororo dataset.

In addition, when comparing configurations with and without the visual agent, those that included the visual agent generally outperformed those without it across almost all metrics, except for BERTScore. In fact, removing the visual agent led to a substantial drop in accuracy of 4.72\%. This indicates that visual cues were particularly dominant for this dataset and highlights the strong contribution of the visual agent. In terms of BERTScore, only the Full configuration outperformed the Language-only and Language + Critic configurations. However, the differences in BERTScore across all configurations were relatively small. 

Overall, the results indicate that the contributions of individual agents vary across datasets, reflecting differences in visual complexity and contextual richness. On the Pororo dataset, eliminating the critic agent causes the most pronounced decline among the tested configurations, suggesting that answer verification and correction are especially important when visual cues are sparse or ambiguous. Removing the visual agent also results in reduced performance, indicating that visual information remains beneficial. On the Simpsons dataset, removing the visual agent results in a clear performance decrease. This suggests that visual descriptions provide strong contextual support for answer generation. Overall, the ablation results highlight how different agents contribute under varying visual and contextual conditions. This underscores the importance of adapting agent design and coordination strategies to the characteristics of specific cartoon-based VQA datasets.

\subsection{Visual Encoder Comparison}
Table~\ref{tab:blip2-comparison} compares the performance of the proposed visual agent with BLIP2 across both datasets. Across both Pororo and Simpsons, replacing the GPT-4o-mini-based visual agent with BLIP2 consistently leads to lower accuracy for both combined settings. This pattern indicates that the generic visual features and captions produced by BLIP2 are less effective than the tailored, cartoon-specific descriptions generated by the proposed visual agent.

A plausible explanation for this gap is that BLIP2 is primarily optimised for natural-image understanding and tends to emphasise object-level semantics, while often overlooking symbolic cues and narrative conventions that are crucial in cartoon imagery, such as exaggerated facial expressions, onomatopoeic text, and stylised visual metaphors. In contrast, the proposed visual agent is explicitly prompted to capture these cartoon-specific elements, providing the downstream language and critic agents with richer and more relevant visual context for answering cartoon VQA queries.

\begin{table}[ht]
    \centering
    \caption{Comparison between BLIP2 visual encoder and GPT-4o-mini-based visual agent (average accuracy, 0--1). Only combined configurations are reported.}
    \label{tab:blip2-comparison}
    \small
    \begin{tabularx}{\linewidth}{@{}Xcc@{}}
        \toprule
        \textbf{Configuration} & \textbf{Pororo} & \textbf{Simpsons} \\
        \midrule
        BLIP2 Visual + Language           & 0.7438 & 0.8000 \\
        GPT-4o-mini Visual + Language          & 0.8313 & 0.8819 \\
        BLIP2 Visual + Language + Critic  & 0.7569 & 0.8056 \\
        GPT-4o-mini Visual + Language + Critic & 0.8375 & 0.8819 \\
        \bottomrule
    \end{tabularx}
\end{table}

\section{Conclusion}
This work investigates multi-agent reasoning with large language models in the context of cartoon-based visual question answering. By analysing the roles of visual, language, and critic agents through systematic ablations, the results indicate that the agents contribute in complementary ways, with the visual agent being most effective for questions grounded in direct visual evidence, the language agent performing better when questions depend on dialogue, narrative context, or commonsense reasoning, and the critic agent improving overall robustness by identifying inconsistencies and reducing hallucinated answers. Multi-agent reasoning is particularly effective when cartoon VQA requires multi-step integration of ambiguous stylised visual cues and narrative context. This effectiveness arises from the decomposition of perception and contextual inference across specialised agents, while feedback from the critic agent enables iterative self-correction, leading to more consistent and grounded answers than a single-pass model. However, introducing multiple agents can also introduce noise and instability, including agent-level variability. Inaccurate visual descriptions or overly conservative critic decisions may propagate errors across agents and, in some cases, lead to worse performance than a well-optimised single-agent baseline when representations are unreliable. Overall, these findings highlight the value of decomposing reasoning into specialised agents for analysing stylised visual narratives, and inform the design of multi-agent architectures for cartoon-based VQA.

\section{Limitations}
Despite providing insights into multi-agent reasoning behaviour, this work has some limitations. The analysis is limited to two cartoon-based datasets, which differ in visual style and narrative structure but do not fully capture the diversity of visual domains. As a result, the observed patterns may not generalise to illustration-based visual domains beyond the cartoon settings considered in this study. In addition, the evaluation relies on automatic metrics and LLM-based judging, which may not fully capture fine-grained reasoning errors. These limitations suggest future work on broader domain evaluation and more evaluation strategies.

\bibliography{reference}

\appendix

\end{document}